\documentclass[runningheads]{llncs}
\usepackage[T1]{fontenc}
\usepackage{graphicx}
\begin{document}
\title{Transformer-based Detection of Multiword Expressions in Flower and Plant Names}
%
%
%
\author{\textsuperscript{1}Damith Premasiri \and
  \textsuperscript{2}Amal Haddad Haddad \and \textsuperscript{1}Tharindu Ranasinghe \and \textsuperscript{1}Ruslan Mitkov}
\authorrunning{D. Premasiri et al.}
%
\institute{\textsuperscript{1}University of Wolverhampton, UK \\
\textsuperscript{2}University of Granada, Spain
\email{\{damith.premasiri,tharindu.ranasinghe,R.Mitkov\}@wlv.ac.uk} \\
{amalhaddad@ugr.es}} 
\maketitle              
\begin{abstract}

Multiword expression (MWE) is a sequence of words which collectively present a meaning which is not derived from its individual words. MWEs detection is an important topic in different natural language processing (NLP) applications, including machine translation. Therefore, detecting MWEs in different domains is an important research topic. With this study, we explore state-of-the-art neural transformers in the task of detecting MWEs in flower and plant names. We evaluate different transformer models on a dataset created from Encyclopedia of Plants and Flower. We empirically show that neural transformers could outperform previous neural models like long-short term memory (LSTM) even in specific domains such as flower and plant names.

\keywords{Multiword Expressions  \and Transformers \and Deep Learning \and Flowers \and Plants}
\end{abstract}
\section{Introduction}
The correct interpretation of Multiword Expressions (MWEs) is crucial to many natural language processing (NLP) applications but is challenging and complex. In recent years, the computational treatment of MWEs has received considerable attention, but there is much more to be done before one can claim that NLP and Machine Translation (MT) systems process MWEs successfully \cite{multiwordunits}.

The study of multiword expressions in NLP has been gaining prominence, and in recent years the number of researchers and projects focusing on them has increased dramatically. The successful computational treatment of MWEs is essential for NLP, including MT and Translation Technology. The inability to detect MWEs automatically may result in incorrect (and even unfortunate) automatic translations and may jeopardise the performance of applications such as text summarisation and web search. 

Multiword expressions do not only play a crucial role in the computational treatment of natural languages. Often terms are multiword expressions (and not single words), making them highly relevant to terminology. The requirement for correct rendering of MWEs in translation and interpretation highlights their importance in these fields. Given the pervasive nature of MWEs, they play a crucial role in the work of lexicographers who study and describe both words and MWEs. Lastly, MWEs are vital in the study of language, which includes not only language learning, teaching and assessment but also more theoretical linguistic disciplines such as pragmatics, cognitive linguistics and construction grammar, which are nowadays aided by (and, in fact, often driven by) corpora. MWEs are very relevant for corpus linguists, too. As a result, MWEs provide an excellent basis for interdisciplinary research and for collaboration between researchers across different areas of study, which for the time being, is underexplored. 

This study is concerned with developing and evaluating a methodology designed to identify multiword expressions among flower and plant names. Multiword expressions are common among the names of flowers and plants, as in the case of \textit{Leontopodium alpinum, White Moonlight or Pink Shirley Alliance}. To the best of our knowledge, this is the first study covering this domain. 

The rest of the paper is structured as follows. Section 2 outlines related work. Section 3 describes the dataset used for our experiments, while section 4 presents the methodology. Section 5 reports the evaluation results, and finally, section 6 summarises the conclusion of this study.

\section{Related Work}
\label{sec:related_work}

Neural models are increasingly employed to detect MEWs. Since both MWEs detection and named entity recognition (NER) tasks are about token classification, they can be modelled using similar models. Therefore we are using a set of models which are used in NER for the MWE detection task, too \cite{rohanian-etal-2019-bridging}.


LSTMs \cite{10.1162/neco.1997.9.8.1735} and gated recurrent units (GRUs) \cite{rohanian-etal-2019-bridging} are the most popular deep learning methods which have been employed in MWEs detection task. There are methods where an LSTM network \cite{10.1162/neco.1997.9.8.1735} is combined with a Conditional random field (CRF) for the same task. Furthermore, graph convolutional neural networks (GCNs) \cite{kipf2016semi} have also been used for MWE identification. The performance of GCNs in MWEs detection has improved using multi-head self-attention \cite{rohanian-etal-2019-bridging}. Transformers have also been used in MWEs detection; \cite{taslimipoor2020mtlb,walshbert}; however, the research has been minimal and has focussed only on general domains.
More specifically, there is limited research on MWEs detection tasks in specific fields, such as MWEs detection in flower and plant names. This paper tries to understand how the state-of-the-art transformer models work in MWEs detection in flower and plant names by empirically evaluating several transformer models.  

\section{Data}
The subject of flowers and plants is of great interest to both professionals and the general public and is relevant to professionals in botany, phytotherapy, plant pharmacology, designers, etc. In addition, there is a lot of interest among the general public as many people dedicate their time to planting plants and growing them in their gardens and homes. Apart from that, the identification of names of flowers and plants as terms is also relevant to terminologists and translators. The study of the names of plants as terms helps in laying the basis of term coining processes and gives insights into the underlying mechanisms of term creation. Translators also benefit from this information for its transfer between languages. For this reason, the automatic identification of names of flowers and plants is relevant to meeting all those needs.

The Encyclopaedia of Plants and Flowers\cite{flowerencyclopedia} of The American Horticultural Society, edited by Christofer Brickel and published by Dorling Kindersley editorial, is used as the corpus for this study. This edition is available in a digitalised format in the online library of the Internet Archive.
This encyclopaedia consists of 522,707 words. It contains a dictionary of names of flowers from around the world, with approximately 8000 terms referring to both scientific and common names and their origins, as well as 4000 images. It also contains descriptions of each flower, instructions on how to plant it and how to use the plants to design gardens.
The dataset has been created by extracting the text from Encyclopedia of Plants and Flowers\cite{flowerencyclopedia}. This encyclopedia consists of two main parts: the plant catalogue and the plant dictionary. The book describes the origins of plant names and basic gardening concepts apart from its main concerns. 
The plant catalogue describes different categories such as Trees, Shrubs, Roses, Perennials etc. In the remaining parts, the plants are subdivided into large, medium and small sub-divisions Ex: large trees, and small trees. The plant dictionary is a dictionary which seeks to cover all possible plant names along with a short description and references to different sections in the plant catalogue accordingly. The flower and plant names in the dictionary are abbreviated similarly to an ordinary dictionary. The data was pre-processed by annotating the terms of proper names. 
The training and test data were created by combining both the plant catalogue and plant dictionary and pre-processing them and tagging the MWE according to the IOB format. The I, O, and B tags stand for Inside, Outside and Beginning, respectively, based on the MWE tag position of the word.
The training set consists of 38,985 sentences, whilst the test set consists of 14,234 sentences.

\section{Methodology}

Transformers based models have produced state-of-the-art art results in many NLP tasks such as text classification \cite{uyangodage-etal-2021-multilingual,ranasinghe-zampieri-2020-multilingual,uyangodage-etal-2021-transformers}, sense disambiguation \cite{hettiarachchi-ranasinghe-2021-transwic,hettiarachchi-ranasinghe-2020-brums}, question answering\cite{damith2022DTWquranqa}, machine translation\cite{wang-etal-2019-learning-deep,ranasinghe-etal-2021-exploratory} and named entity recognition\cite{ranasinghe-etal-2021-wlv,ranasinghe-zampieri-2021-mudes}. Therefore, we employ transformers based models for MWEs detection task while comparing performance with Bidirectional LSTM model. We further discuss these models in the following sections. 

\begin{figure}[ht]
\centering
\includegraphics[scale=0.5]{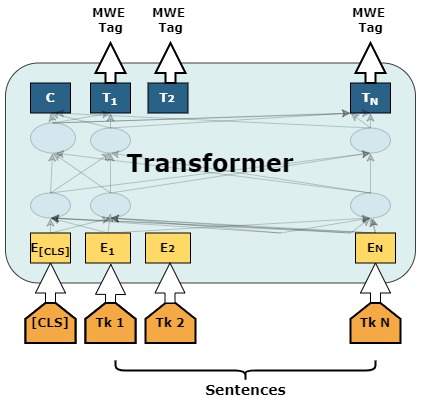}
\caption{MWEs token classification transformer architecture \cite{Premasiri2022}}
\label{fig:architecture}
\end{figure}

\paragraph{Transformer} models such as BERT \cite{devlin-etal-2019-bert} have been trained using masked language modelling objective and they can be fine-tuned for multiple different tasks \cite{alloatti-etal-2019-real}. Within this study we fine-tune number of transformer models for MWEs detection which is a token classification task. We modify the original BERT architecture by adding a token level classifier following the last hidden layer as shown in Figure \ref{fig:architecture} to achieve the architecture of the MWE detection model. 
This is a linear classification layer which uses the last hidden state as input and output the relevant token per word such as B,I and O.


\noindent We conducted our experiments with many popular transformer models to detect MWEs such as BERT \cite{devlin-etal-2019-bert}, RoBERTa \cite{zhuang-etal-2021-robustly}, XLNet \cite{NEURIPS2019_dc6a7e65}, XLM-RoBERTa \cite{conneau-etal-2020-unsupervised} and Electra \cite{clark2020electra}. We evaluated several different BERT variations such as bert-base-cased, bert-base-uncased, bert-base-multilingual-cased,  bert-base-multilingual-uncased. Furthermore we experimented sci-bert-cased\cite{Beltagy2019SciBERT} and sci-bert-uncased\cite{Beltagy2019SciBERT} which are pre-trained on scientific corpus as more specific variants of BERT\cite{devlin-etal-2019-bert}. The other transformer models are evaluated only on their base model. All the transformer-based methods made use of a batch size 32, Adam optimiser with learning rate 4e-5. They were trained for 3 epochs with linear learning rate warm-up over 10\% of the training data. These experiments were carried out in an NVIDIA GeForce RTX 2070 GPU and in Google colab GPUs\footnote{https://colab.research.google.com/}.

\paragraph{BiLSTM-CRF} is another token classification architecture which provided state-of-the-art results before transformers \cite{Huang2015BidirectionalLM} . Bidirectional LSTM (BiLSTM) is an improved version of conventional LSTMs which is capable of learning contextual information both forwards and backwards in time. Unlike standard LSTM, the input flows in both directions, and it’s capable of utilizing information from both sides. It utilises an additional LSTM layer which reverses direction of the information flow.  This study leverages the BiLSTM architecture given its' bidirectional ability to model temporal dependencies. CRFs \cite{10.5555/645530.655813} are a statistical model that are capable of incorporating context information and are highly
used for sequence labelling tasks. 
The BiLSTM-CRF model provides a way of combining the relationships of consecutive outputs of the network and utilise both past and future tag information to for prediction. Since the BiLSTM has both of the past and future information, combining CRF on top of the BiLSTM provides powerful network for precisely predicting the tags. BiLSTM-CRF experiments were performed on a CPU with a learning rate of 1e-3 and the model was trained for 60 epochs.

\section{Results}
In this section, we report the results of conducted experiments using macro averaged F1 as it is widly used in classification tasks. As shown in Table \ref{tab:results_table} it is clear that the transformer-based models outperform the BiLSTM-CRF method with clear margins. The BiLSTM-CRF could achieve only 0.3320 Macro F1 score, while all the transformer models we experimented outperformed that. A noticeable observation is that best transformer model had nearly double the F1 score of BiLSTM-CRF model while all the other transformer models performed competitively.

\begin{table}[ht]
    \centering
    \resizebox{8cm}{!}{
        \begin{tabular}{c|c}
            \bf Model & \bf{Macro F1} \\
            \hline
            roberta-base & 0.5039 \\ 
            xlm-roberta-base & 0.5650 \\ 
            xlnet-base-cased & 0.6312 \\ 
            bert-base-multilingual-cased & 0.6086 \\ 
            bert-base-multilingual-uncased & \bf0.6422 \\ 
            bert-base-cased & 0.6393 \\ 
            bert-base-uncased & 0.6227 \\ 
            electra-base-discriminator & 0.5753  \\ 
            sci-bert-cased & 0.6214  \\ 
            sci-bert-uncased & 0.6307  \\ 
            \hline
            BiLSTM-CRF & 0.3320 \\
        \end{tabular}
    }
    \caption{Results for multiword expression detection in flower and plant names}
    \label{tab:results_table}
\end{table}

The clear winner is the bert-base-multilingual-uncased model with a Macro F1 score of 0.6422. This is followed by bert-base-cased and xlnet-base-cased models with Macro F1 scores of 0.6393 and 0.6312, respectively. In general, transformer models have better performance with slight margins among them. An interesting observation is that multilingual-bert model outperforms sci-bert models, which are trained on a corpus featuring a high frequency of scientific terms. We conjecture that this could be due to a lack of the flower names and plant names related data in the sci-bert training set. Nevertheless, sci-bert-uncased model competitively performed with 0.6307 Macro F1 score, which is only a 0.0115 difference from the best performing model.  

Another interesting observation was while the multi-lingual bert model was the best performer, the cross-lingual model; xlm-roberta-base did not do well with a F1 score of 0.5650. This score is very close to the least successful model among transformers which was roberta-base with Macro F1 of 0.5039. Yet these values are outperforming the BiLSTM-CRF model, which shows the powerful nature of transformers based models in MWE tasks over the other neural methods like BiLSTM. 

Overall, transformers-based neural methods clearly perform better than BiLSTM-CRF. All the transformer-based methods performed above 0.5000 of Macro F1, showing their strong performance in MWE detection tasks. 




\section{Conclusion}
MWE detection has significant importance in many NLP applications, especially in translation and terminology studies. In this paper, we focus on an empirical analysis of multiple neural transformer models in the MWE detection task using a flowers and plants dataset. We show that all transformer models outperform the LSTM-based method. Of the transformer models we experimented with, bert-base-multilingual-uncased reported the best results doing better than other transformer models. We can conclude that transformer models can handle the challenges presented by MWEs in local domains like plant names and flower names better than the previous neural methods, such as LSTM.


In the future, we would like to explore more specific domains similar to flower and plant names. It would be interesting to study how MWEs detection works in different languages with different flower and plant names. We are encouraged to explore cross-lingual models more in this regard to understand how well these models perform across languages on the MWEs detection task for similar datasets.

%
%
%
\bibliographystyle{splncs04}
\bibliography{bibliography}





\end{document}